\documentclass[pdftex]{bmvc2k}

\usepackage{times}
\usepackage{amsmath}
\usepackage{amssymb}

\usepackage{graphicx}
\usepackage{color}
\usepackage{amsfonts}
\usepackage{subfigure}

\newcommand{\fig}[1]{{Figure~\ref{#1}}}

\newcommand{\secn}[1]{{Section~\ref{#1}}}

\newcommand{\argmin}{\mathop{\rm arg~min}\limits}

\newcommand{\secspace}{{\ \ \ \ \ }}

\makeatletter
\newcommand{\figcaption}[1]{\def\@captype{figure}\caption{#1}}
\newcommand{\tblcaption}[1]{\def\@captype{table}\caption{#1}}
\makeatother

\title{Physical Cue based Depth-Sensing by Color Coding with Deaberration Network}

\addauthor{Nao Mishima}{nao.mishima@toshiba.co.jp}{1}
\addauthor{Tatsuo Kozakaya}{tatsuo.kozakaya@toshiba.co.jp}{1}
\addauthor{Akihisa Moriya}{akihisa.moriya@toshiba.co.jp}{1}
\addauthor{Ryuzo Okada}{ryuzo.okada@toshiba.co.jp}{1}
\addauthor{Shinsaku Hiura}{hiura@eng.u-hyogo.ac.jp}{2}

\addinstitution{Corporate Research and Development Center, Toshiba Corp.}
\addinstitution{University of Hyogo}

\runninghead{Mishima ET AL.}{Depth Sensing with Deaberration Network}

\begin{document}


\maketitle

\begin{abstract}
Color-coded aperture (CCA) methods can physically measure the depth of a scene given by physical cues from a single-shot image of a monocular camera. However, they are vulnerable to actual lens aberrations in real scenes because they assume an ideal lens for simplifying algorithms. In this paper, we propose physical cue-based deep learning for CCA photography. To address actual lens aberrations, we developed a deep deaberration network (DDN) that is additionally equipped with a self-attention mechanism of position and color channels to efficiently learn the lens aberration. Furthermore, a new Bayes L1 loss function based on Bayesian deep learning enables to handle the uncertainty of depth estimation more accurately. Quantitative and qualitative comparisons demonstrate that our method is superior to conventional methods including real outdoor scenes. Furthermore, compared to a long-baseline stereo camera, the proposed method provides an error-free depth map at close range, as there is no blind spot between the left and right cameras.
\end{abstract}


\section{Introduction}

\begin{figure}
\centering
  \includegraphics[width=8cm]{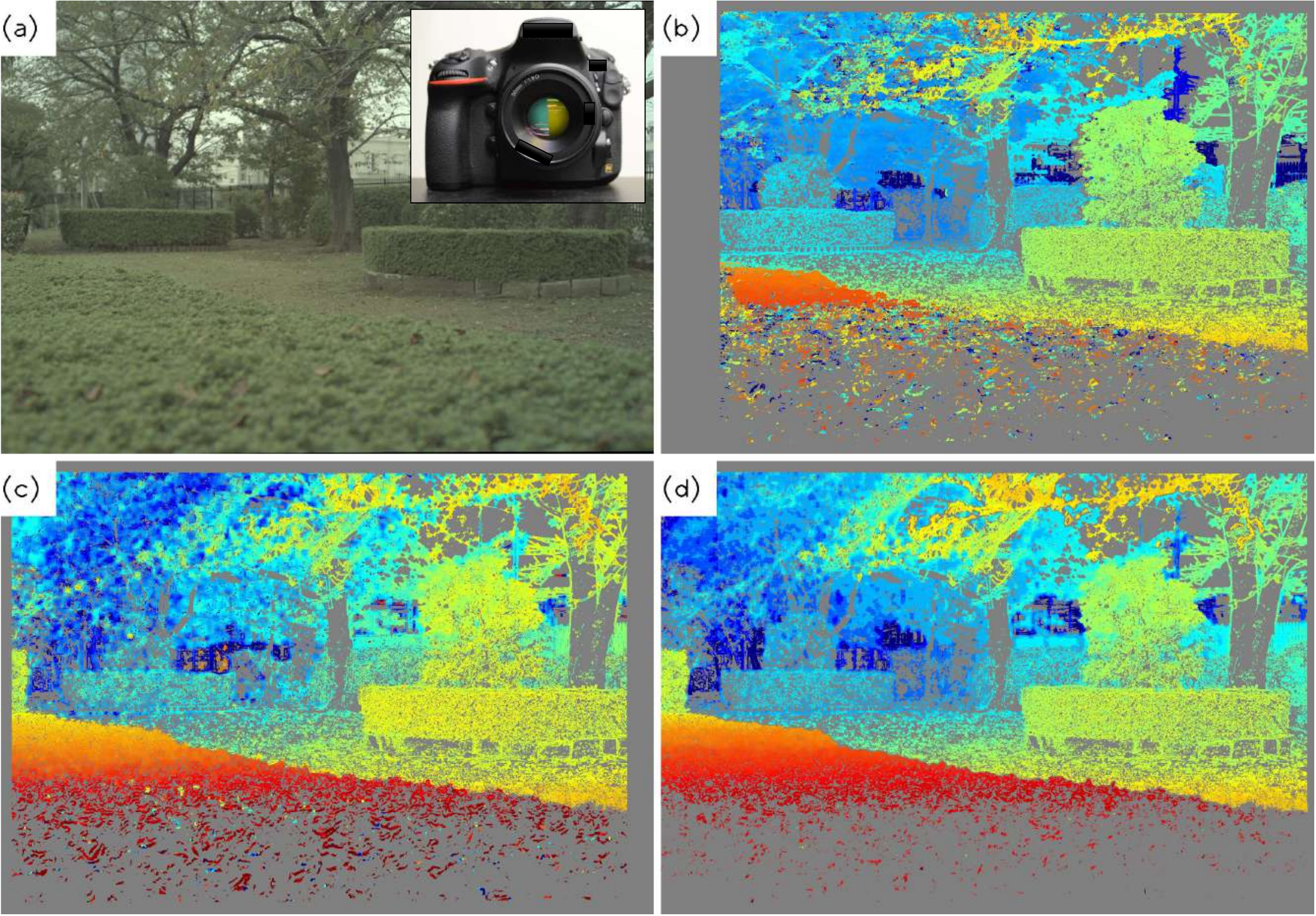}
  \figcaption{(a)Our prototype camera with color coded aperture and its captured image. (b)Depth map estimated by a stereo camera for the reference. (c)Depth from the analytical defocus (DfAD)~\cite{moriuchi201723} (d)Depth obtained by the proposed method with a deep deaberration network (DDN). In the distant regions, our method is superior to the conventional one.}
  \label{title_figure}
\end{figure}

Compared with multi-shot depth measurement methods such as structure from motion (SfM)~\cite{agarwal2009building} and depth from defocus (DfD)~\cite{subbarao1994depth,tang2017depth,guo2017focal}, a single-shot method is suitable for moving objects. One of the most successful single-shot methods is deep monocular depth estimation. Despite the remarkable progress of deep monocular depth estimation in recent years~\cite{eigen2014depth,godard2017unsupervised,fu2018deep}, it cannot estimate a correct depth map without sufficient contextual information due to the lack of a physical depth cue, for instance, in a scene without the ground. 

\begin{figure}
    \centering
    \includegraphics[width=9cm]{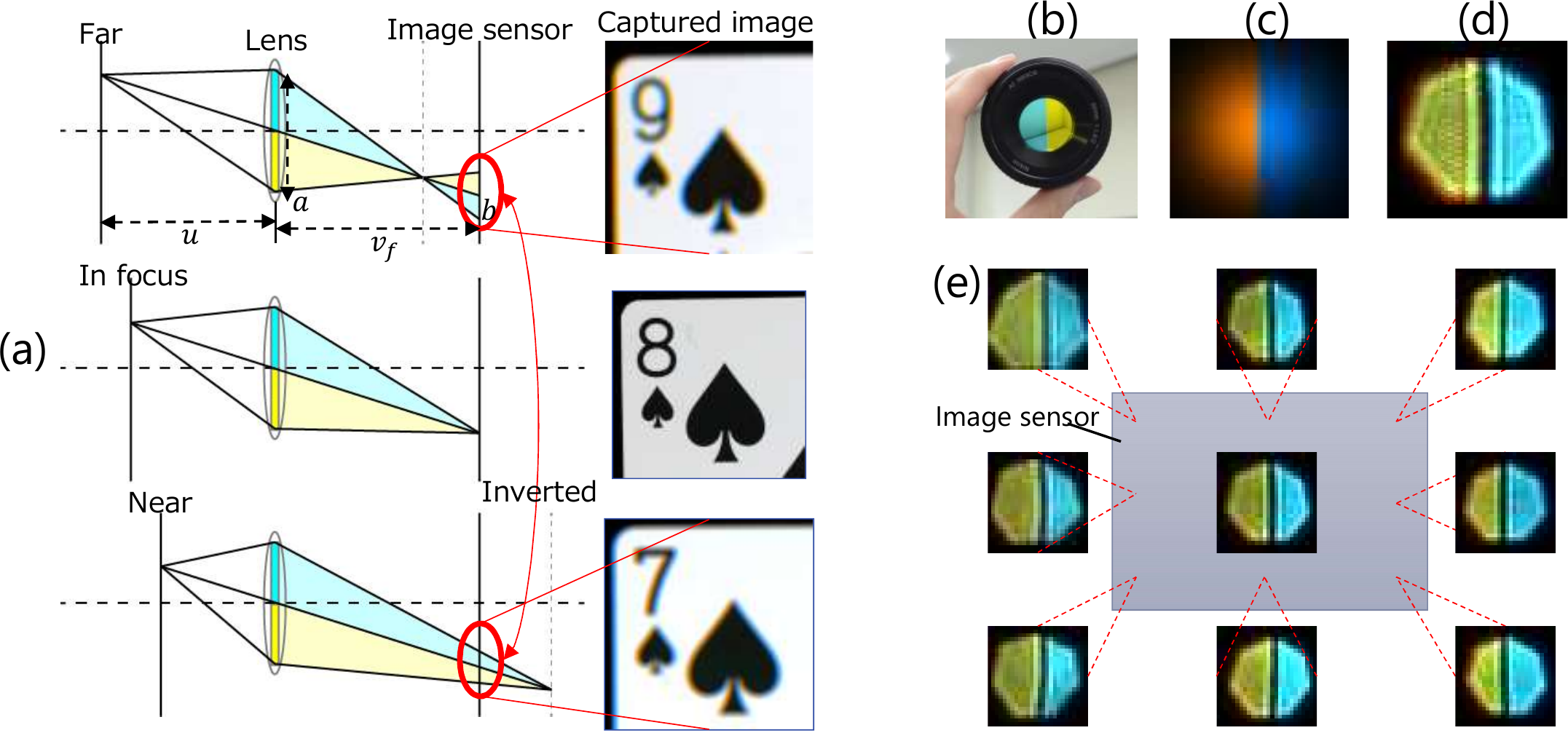}
    \caption{(a)Depth cue in the image captured with CCA. (b)Our prototype CCA lens (c)An analytical PSF assumed by a conventional CCA method\cite{moriuchi201723} (d)An actual PSF at the center of the image (e)Shift variance of PSF on an actual lens.}
  \label{psf_aberration}
\end{figure}

Instead of utilizing contextual information, color-coded aperture (CCA) methods can acquire a depth map based on a physical depth cue encoded in a single-shot image by inserting different types of color filters~\cite{bando2008extracting,kim2012multifocusing,chakrabarti2012depth,lee2013single,martinello2015dual,paramonov2016depth,moriuchi201723} into the lens aperture as shown in \fig{psf_aberration}b. As shown in \fig{psf_aberration}a, color and blur radius vary according to the distance from the focus distance. Conventional CCA methods assume an ideal lens for simplifying analytical modeling of defocus blur. However, actual lenses have shift-variant point spread functions (PSFs) distorted according to the position of the image sensor by lens aberrations such as field curvature, coma or lateral chromatic aberration as shown in \fig{psf_aberration}e. Furthermore, the depth cues often disappear because of several uncertainties, such as saturation, soft shadow, dark color and large blur. These uncertainties are not treated distinctly in conventional methods~\cite{bando2008extracting,kim2012multifocusing,chakrabarti2012depth,lee2013single,martinello2015dual,paramonov2016depth,moriuchi201723,haim2018depth}. 

In this paper, we propose a physical cue-based deep learning to overcome the differences between the analytical model and the actual one. In order to estimate a correct depth map under shift-variant PSFs, we add positional information as an additional branch by a self-attention mechanism~\cite{zhang2018self}. It also couples additional color channels to solve dependency on object colors for handling various complex color pattern correctly. To handle various uncertainties, we improve the loss function based on Bayesian deep learning~\cite{kendall2017uncertainties} for stabilizing the training. As shown in \fig{title_figure}, We demonstrate that our method is superior to a conventional method in quantitative and qualitative experiments including various outdoor scenes. Furthermore, compared to a long-baseline stereo camera, the proposed method provides an error-free depth map at close range, as there is no blind spot between the left and right cameras. 

The contributions of this paper are as follows. {\bf (a)} We propose a CNN-based depth estimation network that does not infer the depth from the contextual information but physically measures the depth given by an optical cue. {\bf (b)} We add positional information as additional channels by a self-attention mechanism to handle shift-variant aberrations. {\bf (c)} We train the network with additional color channels using many pictures taken by an actual lens to handle various complex color patterns correctly. {\bf (d)} To handle various uncertainties, we propose Bayes L1 loss instead of the conventional heteroscedastic variance for stabilizing the training. 
There are two limitations to this paper. First, our method is not applicable to the regions with small gradients since there are no depth cues. Second, our method requires much computation time, for example, about 50 seconds (NVIDIA Geforce GTX 1080 ti), because of the patch-based architecture~\cite{bailer2018fast}. The improvement is left for future work. 

\section{Background(Color-coded aperture photography)}\label{sc_CCA}


CCA methods~\cite{bando2008extracting,kim2012multifocusing,chakrabarti2012depth,lee2013single,martinello2015dual,paramonov2016depth,moriuchi201723} are categorized to a computational photography (CP) technique~\cite{zhou2011computational} developed in the last decade. The image quality of CCA is higher than coded-aperture~\cite{levin2007image} having unnatural blur shape due to the special shape of the aperture. In order to acquire the depth map, CCAs use disparity~\cite{bando2008extracting,kim2012multifocusing,lee2013single,paramonov2016depth} or defocus blur~\cite{chakrabarti2012depth,martinello2015dual,moriuchi201723}.

\begin{figure}
\centering
  \includegraphics[width=13cm,bb=0 0 2486 828]{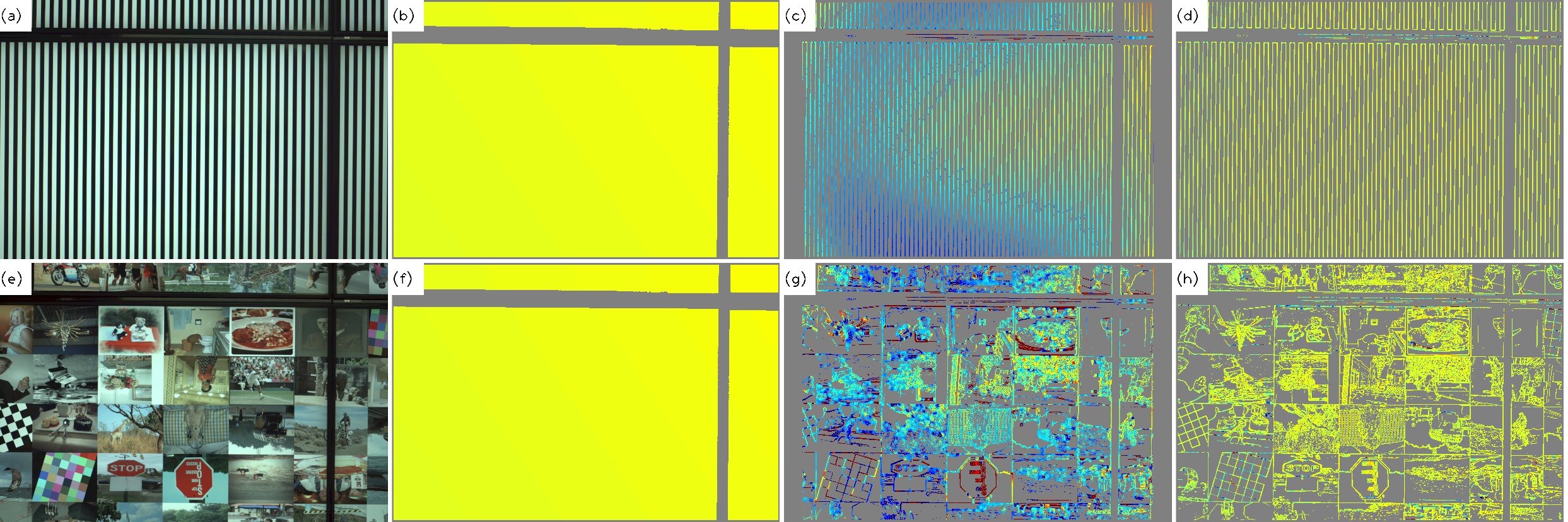}
  \caption{(a), (e) Captured images. (b), (f) Ground truths. (c), (g) DfAD\cite{moriuchi201723}.  (d), (h) DDN(ours). }
  \label{depth_map_system}
\end{figure}

\fig{psf_aberration}a shows the change of the optical path through the lens with cyan and yellow color filters~\cite{paramonov2016depth,moriuchi201723}. The color direction of defocus blur with a near or far object is inverted at the focus distance. Such a change of the defocus blur allows retrieval of the distance in front of or behind the focal plane. 
Depth from defocus technique~\cite{subbarao1994depth} is applied to estimate an accurate depth map for CCA~\cite{martinello2015dual,moriuchi201723}, which is called depth from analytical defocus (DfAD). These methods assume the gaussian blur as shown in \fig{psf_aberration}c. The blur radius is estimated by 
$
\hat{b} = \argmin_{b} 3 - D(\nabla I_R(b),\nabla I_G)-D(\nabla I_G,\nabla I_B(b))-D(\nabla I_R(b),\nabla I_B(b))
$
, where $I_R(b)=k_R(b)*I_R$ and $I_B(b)=k_B(b)*I_B$ are deformed images by convolution kernels $k_R(b), k_B(b)$ deforming the asymmetric gaussian blur of the R and B images to the gaussian blur of the G image and $D$ is zero-mean normalized cross correlation~\cite{moriuchi201723}. 
However, conventional CCA methods assume an ideal lens for simplifying analytical modeling of the defocus blur. By the difference between the analytical model and the actual one, as shown in \fig{depth_map_system}c, DfAD gives a distorted depth map due to a shift-variant PSF(\fig{psf_aberration}d). \fig{depth_map_system}f shows errors because of differences between the ideal analytical model and the actual one. Although recent work~\cite{paramonov2016depth} modeled the aberration effect by a double-Gauss lens model, the handcrafted model must be reconstructed when it is applied to other lenses. 


\section{Method}\label{sc_method}

In this section, to overcome the differences between the analytical model and the actual one, we propose a physical cue-based deaberration network. 

\begin{figure*}[h]
\centering
  \includegraphics[width=11.0cm]{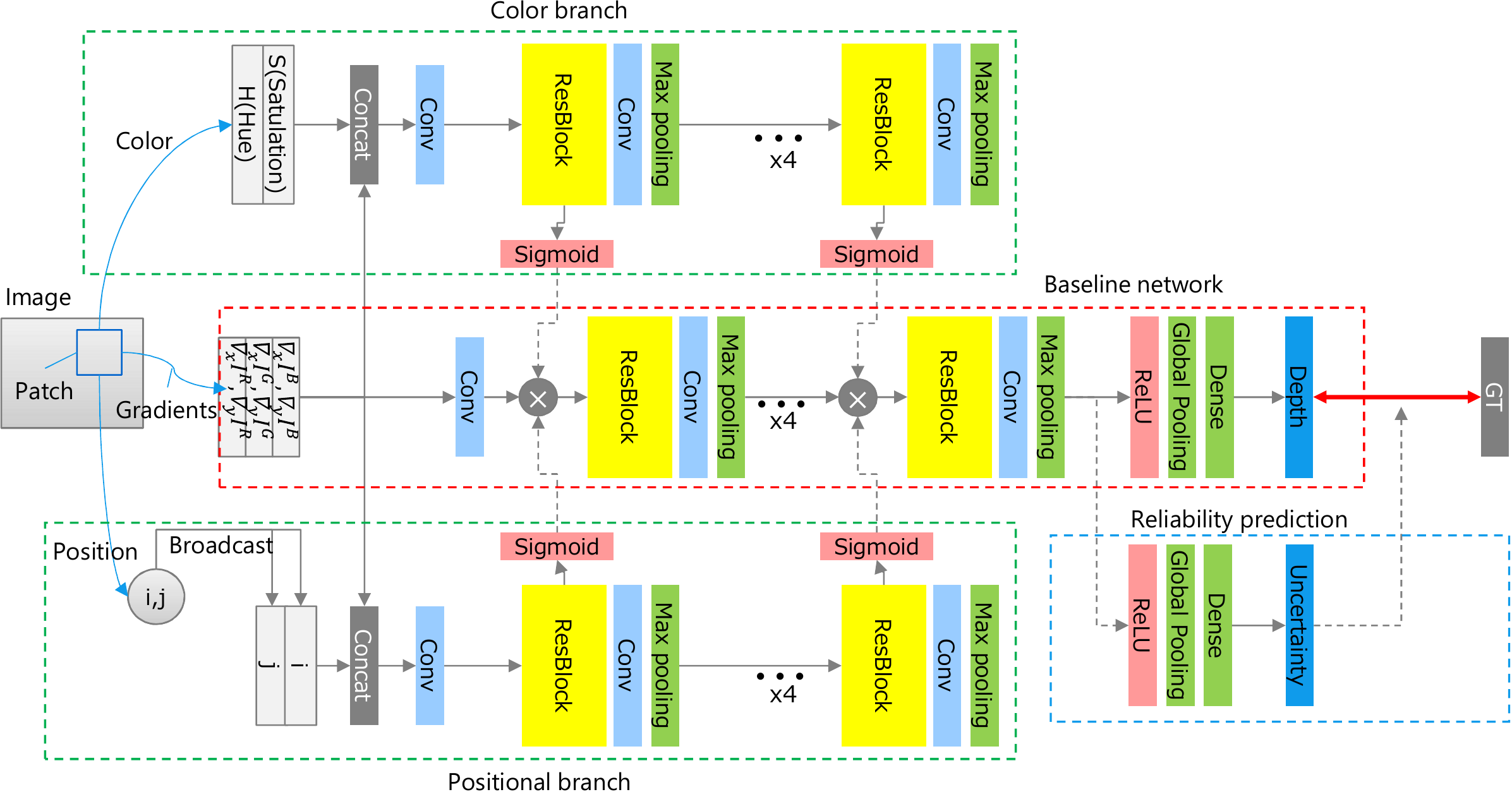}
  \caption{The structure of our deep deaberration network (DDN) based on ResNet. The main branch extracts defocus blur from the gradients. Positional and color branches make attention maps. The feature map of the main branch is multiplied by these attention maps. }
  \label{attention}
\end{figure*}

\noindent{\bf Baseline network}\ \ \ 
We adopt patch-based architecture~\cite{zagoruyko2015learning,simo2015discriminative,zbontar2016stereo,luo2016efficient,bailer2016cnn,bailer2018fast} for our network to learn only the defocus blur instead of the contextual information and train the network easily. The architecture takes a patch extracted from a captured image as an input and outputs a single depth value corresponding to the patch. Since this network does not access to information of neighbor patches, it does not learn the contextual information. Therefore, the learned network has high generalization performance. This network can be trained by patchwise images with flat depth data only. Such data can be collected easily by our training system (see \secn{sc_Image_capture_system}). 

Our network structure is based on ResNet~\cite{he2016deep}. The baseline network structure is indicated by the red dotted rectangle in \fig{attention}. In a preprocessing stage, the gradient of an image patch is calculated with respect to the horizontal and vertical axis. All of the gradients are concatenated to $x(i,j)$. It is well known that gradients give better results than color images do~\cite{bando2008extracting,chakrabarti2012depth,martinello2015dual,paramonov2016depth,moriuchi201723} and our experiment also has shown such result (see also \secn{sc_verifications}). The gradients are processed by several ResBlocks, a convolution layer and max-pooling, followed by an activation function (ReLU) and global average pooling~\cite{lin2013network} and a fully connected layer (dense layer). The network infers the defocus blur with learnable weight parameters $\theta$ as $\hat{b}(i,j) = f(x(i,j);\theta)$. 

\noindent{\bf Deep deaberration network}\secspace
The aberration effect varies according to wavelength of color, horizontal and vertical axis in the lens. In order to handle the aberration effect, we add the positional and color information to the gradients as shown in \fig{attention}. The position $(i, j)$ is broadcasted into the same size as the patch. In order to add color information, the input image patch is converted to hue and saturation with the same shape of the patch. 

In order to handle the lens aberration efficiently, we introduce the self-attention mechanism~\cite{zhang2018self} to our deep deaberration network (DDN) indicated by the green dotted rectangle in \fig{attention}. Since the lens aberration causes the shape of the blur to change, important features vary according to the position of the image patch. The attention mechanism is trained so as to put large weights on such important features accordingly, and, thus, shift-invariant features are extracted as a result. The color branch can handle the dependency on object colors in the same way. 
After concatenating the positional and color information to the gradients, the attention maps are calculated by sigmoid functions from each feature map. The feature map of the main branch is multiplied by the above two attention maps before its ResBlock. 

The proposed networks are trained as a regression problem with supervision similar to stereo matching~\cite{kendall2017end,chang2018pyramid} and deep monocular depth estimation~\cite{kuznietsov2017semi,Atapour-Abarghouei_2018_CVPR}. The ground truth distance $u(i,j)$ recorded by the training system is converted to the blur radius $b(i,j)$ by using the lens maker's formula. A tuple of $(k, x(i,j), u(i,j))$ is the element of a training data-set, where $k \in \{0,...,K-1\}$ is the index. L1 loss function is defined as 
$L(\theta) = \frac{1}{N}\sum_{k}|b(i,j) - f(x(i,j);\theta)|,$
where $N$ is the total number of the training patches. 

\noindent{\bf Reliability prediction}\secspace
In actual CCA optics, the depth cues often disappear because of several uncertainties, such as saturation, soft shadow, dark color, and large blur. In the literature of Bayesian deep learning~\cite{gal2016dropout,kendall2017uncertainties}, such uncertainties are categorized as heteroscedastic uncertainty~\cite{kendall2017uncertainties}. To handle the heteroscedastic uncertainty, the network should be changed to also output variance prediction as $[\hat{b}(i,j), \hat{\sigma}(i,j)] = f(x(i,j); \theta)$. The loss function is defined as heteroscedastic variance~\cite{kendall2017uncertainties}. However, this loss function shows significant instability in the training of our task as shown in \fig{loss_bayes} (indicated by Bayes L2). This instability often causes training to fail. The progress of the training makes the variance prediction noticeably smaller and the error $b'(i,j) - \hat{b}(i,j)$ is also expected to be small. However, outlier errors make the loss very high because the denominator becomes very small simultaneously. Then, the loss function will diverge with the second order. 

To stabilize the training, we propose a new loss function that has the heteroscedastic absolute standard deviation. In order to reduce the order, we convert the loss function by replacing the squared error and the variance with the absolute error and the absolute standard deviation as follows. 
\begin{equation}
L(\theta) = \frac{1}{2N} \sum_{k}\frac{|b'(i,j) - \hat{b}(i,j)|}{{|\hat{\sigma}(i,j)|}} + \log{|\hat{\sigma}(i,j)|}
\label{eq_bnn_l1_loss1}
\end{equation}
To output $|\hat{\sigma}(i,j)|$, an additional final layer is added to the end of the main branch in DDN indicated by the blue dotted rectangle in \fig{attention}. We use $|\hat{\sigma}(i,j)|$ as the reliability. \fig{loss_bayes} shows that our new loss function stabilizes the training significantly (indicated by Bayes L1). 

\section{Experiments}

\begin{figure}
\centering
  \includegraphics[width=11cm]{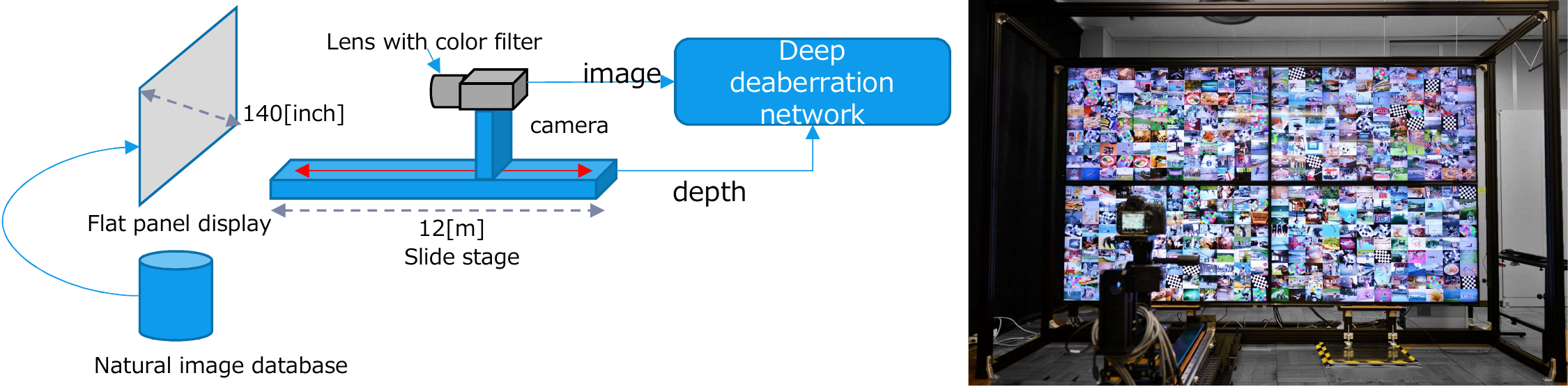}
  \caption{Training data gathering system for DDN. }
  \label{image_capture_system}
\end{figure}

\subsection{Training system, data and details}\label{sc_Image_capture_system}

\noindent{\bf Training system}\secspace
We have developed a training system in order to automatically take many pictures with actual lenses as shown in \fig{image_capture_system}. This system consists of four 8K displays (LC-70X500) and a 12[m] slide stage arranged orthogonally to the displays. As these four displays are arranged 2 x 2, the screen size and resolution become 140 inches and 15360 x 8640, respectively. 
In order to learn only defocus blur information instead of contextual one, we introduce various randomization techniques to our training recipe to make the deep network focus on the blur information. We use many images sampled randomly from the MSCOCO data-set~\cite{vinyals2017show}. They are arranged in a matrix form as shown in \fig{image_capture_system}. Horizontal/vertical flipping and random scaling are applied to each image to remove its shape and scale information. 

\noindent{\bf Training data}\secspace
We used a digital single-lens reflex (DSLR) camera: Nikon AI AF Nikkor 50mm f/1.8D (lens), Nikon D810 (body). The f-number was set to 4.0 throughout all experiments. The focus distance was set to 1500[mm] and the images were taken at 100 positions spaced at regular intervals on the blurred space from 1100[mm] to 2400[mm]. Four different images were taken at each position. Three images were for the training data and the last one was for the test data. This process took about only three hours. The captured images were resized to 1845x1232. We randomly collected image patches from only an edge and texture region without overlapping. The training and test data-sets include around 150,000 and 15,000 patches, respectively. 

\noindent{\bf Implementation and training details}\secspace
Our DDN operates on an input patch size of 16x16 pixels with five Resblocks for each branch. The convolutional layers in all of our networks have 3x3 kernels and 1 stride. The number of channels is fixed to 32 from the beginning to the end. To train our networks, we use ADAM~\cite{kingma2014adam} with the default parameters and 128 as the batch size. Although several data augmentation techniques~\cite{krizhevsky2012imagenet} are usually applied in order to avoid overfitting, these techniques deform the shape of PSF which we should learn. We only select random crop~\cite{buslaev2018albumentations}, brightness~\cite{buslaev2018albumentations} and random erasing~\cite{zhong2017random} that do not affect PSF. We trained DDN by 1500 epochs with the above training data and our training recipe. Finally, the test accuracy reached to 0.72 ($<\sigma$=8.1[mm] at 1500[mm]). 

\subsection{Verifications}\label{sc_verifications}

\begin{figure}
\begin{tabular}{cc}
  \begin{minipage}{.3\textwidth}
  \centering
  \includegraphics[width=3.9cm]{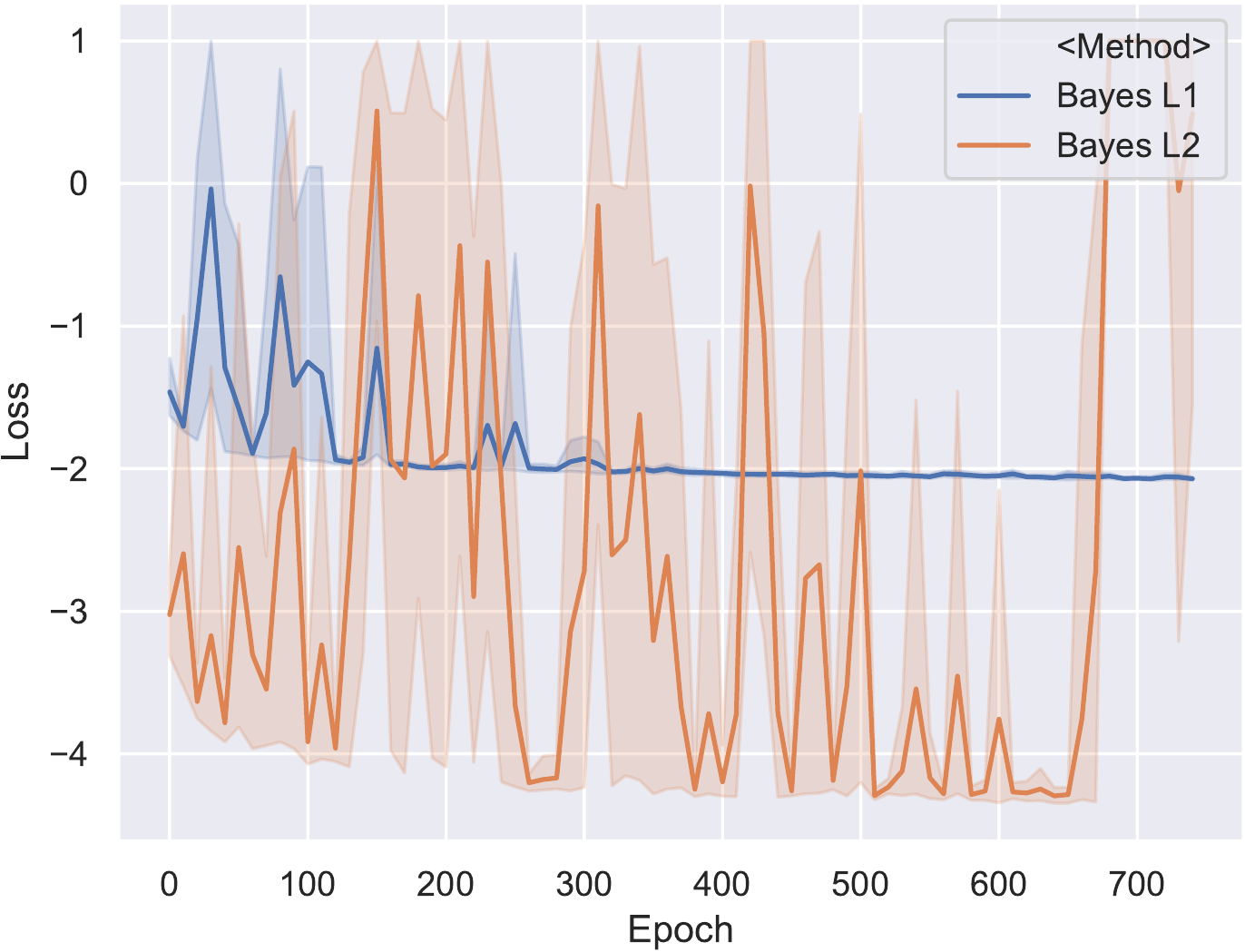}
  \caption{Loss curves during training.}
  \label{loss_bayes}
  \end{minipage}
  \ \ \ \ 
  \begin{minipage}{.3\textwidth}
\centering
  \includegraphics[width=3.9cm]{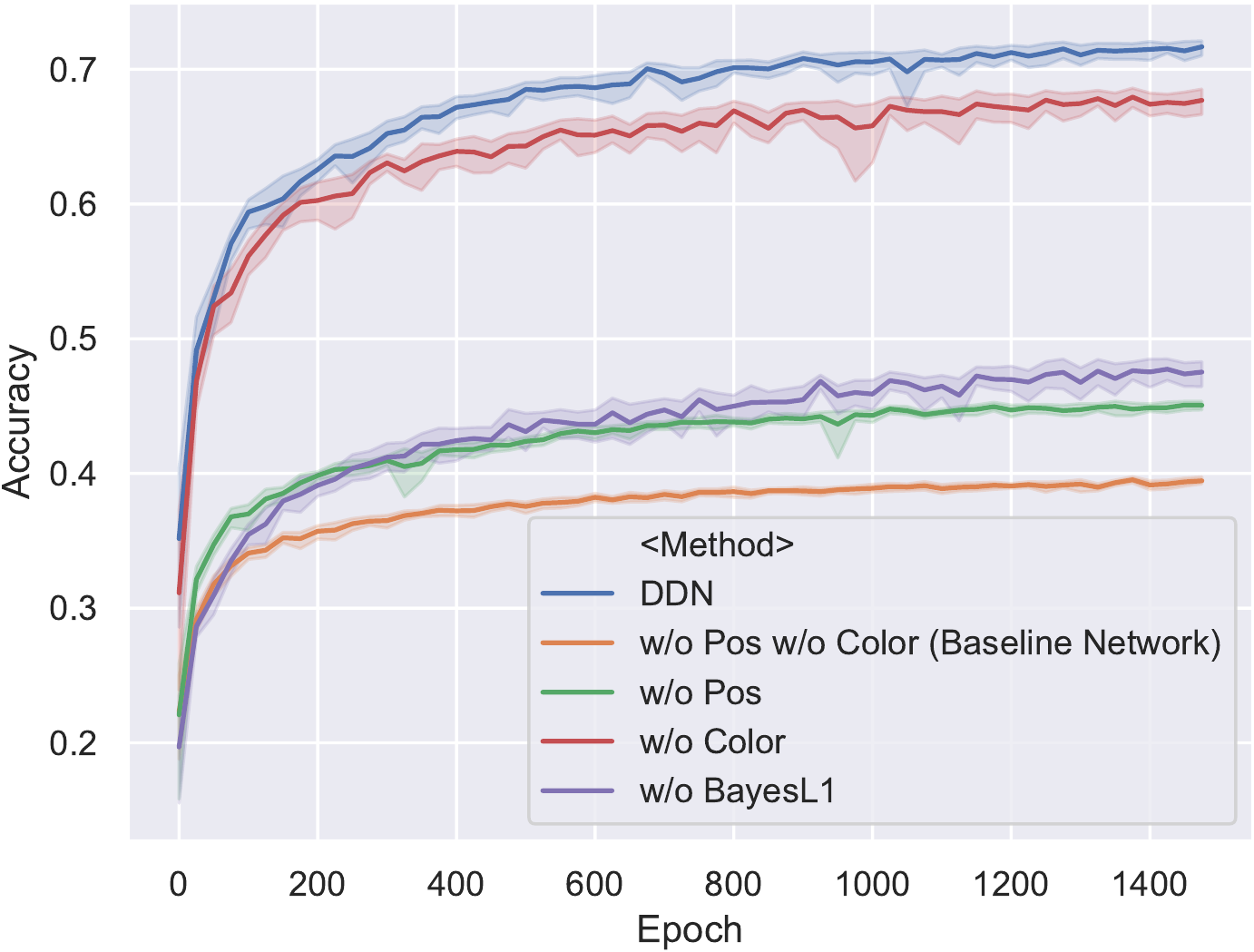}
  \caption{Test accuracy curves for ablation study. }
  \label{ablation}
  \end{minipage}
  \ \ \ \ 
  \begin{minipage}{.3\textwidth}
\centering
  \includegraphics[width=3.9cm]{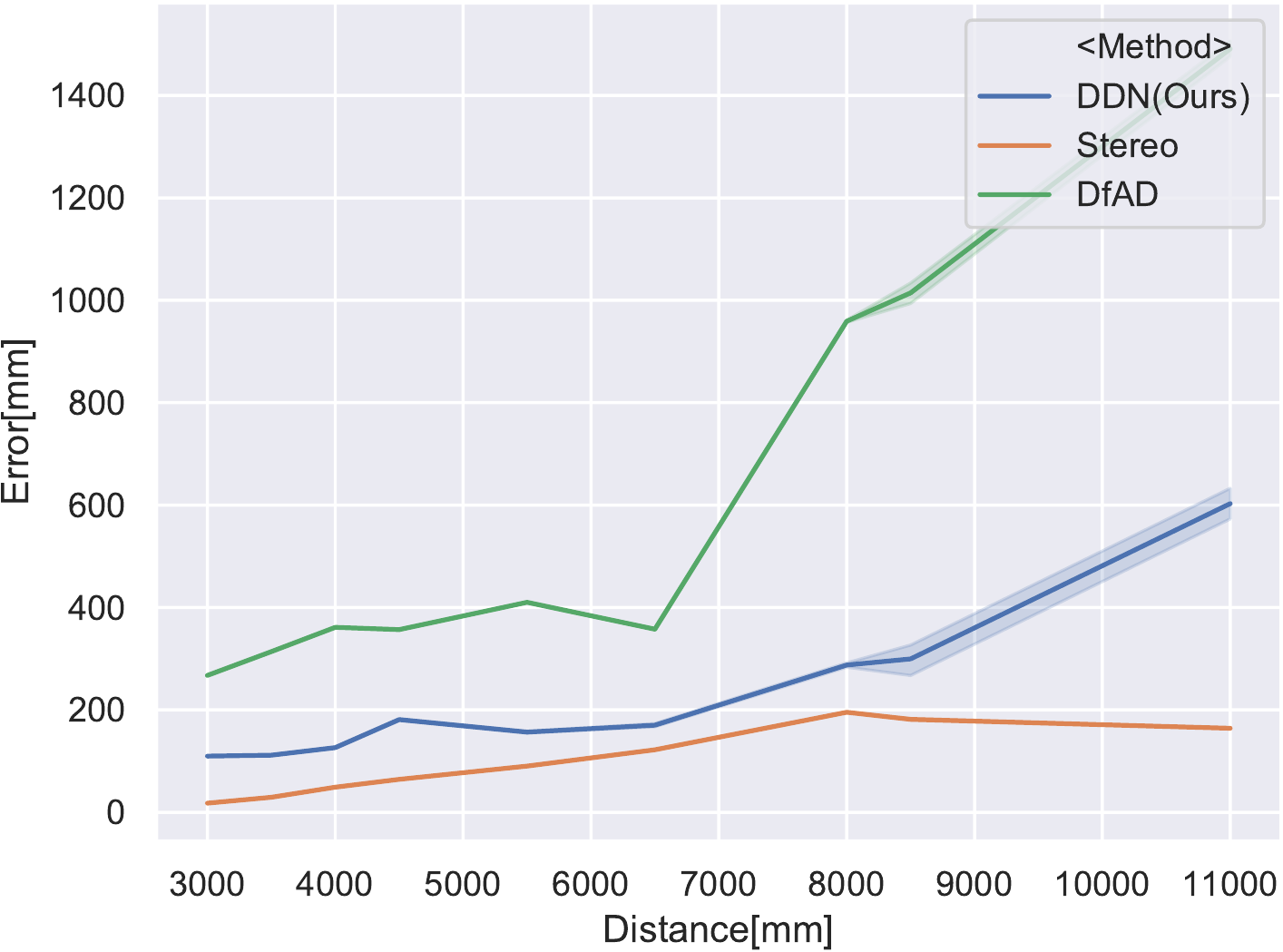}
  \caption{The error curves over target distance. }
  \label{quantitative_result}
  \end{minipage}
\end{tabular}
\end{figure}

\noindent{\bf Ablation study}\secspace
We show the contributions of the proposed components by ablation study. Test accuracy curves during the training for the ablation study is shown in \fig{ablation}. The result shows that the positional branch significantly affects accuracy. The Bayes L1 loss and the color branch have an effect on the accuracy. The gradient affects the accuracy slightly. 

\begin{figure}
\begin{tabular}{cc}
  \begin{minipage}{.47\textwidth}
\centering
  \includegraphics[width=4.5cm]{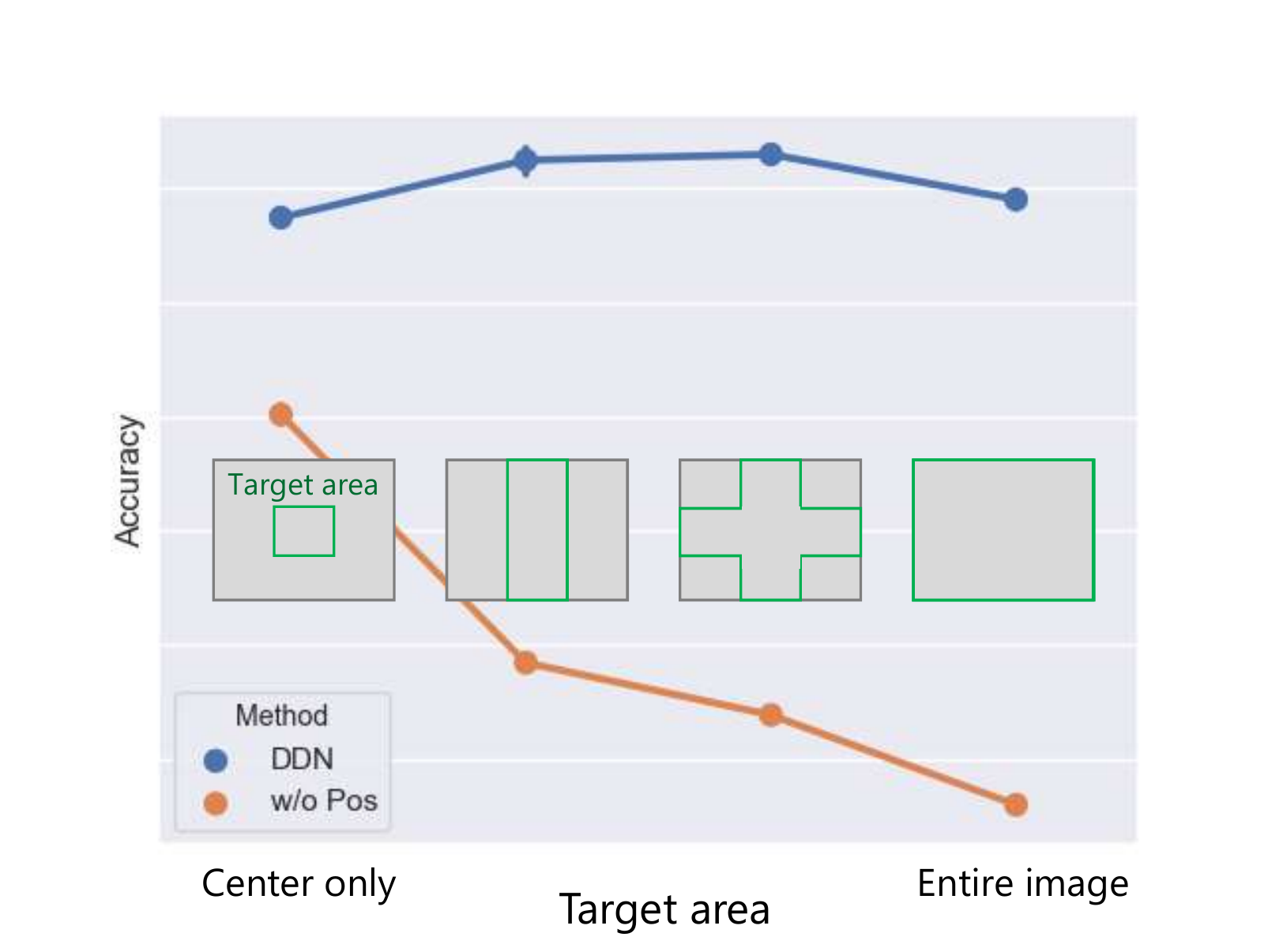}
  \caption{The relationships between the depth estimation area and the accuracy.}
  \label{plot_pos}
  \end{minipage}
  \ \ \ \ \ 
  \begin{minipage}{.47\textwidth}
\centering
  \includegraphics[width=4.5cm]{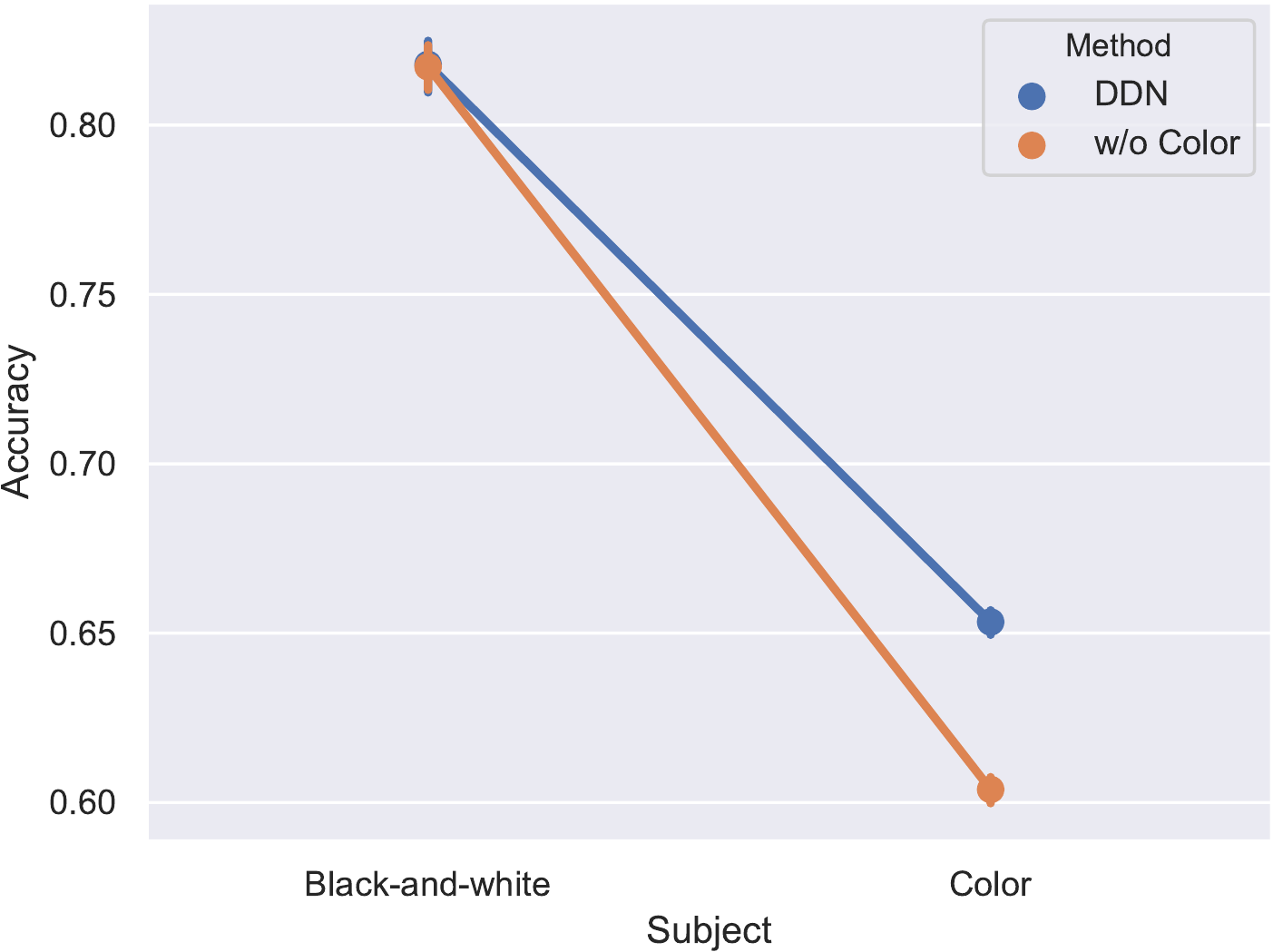}
  \caption{The relationships between the subject's color and the accuracy.}
  \label{plot_color}
  \end{minipage}
\end{tabular}
\end{figure}

\begin{figure}
\centering
  \includegraphics[width=13cm,bb=0 0 2486 828]{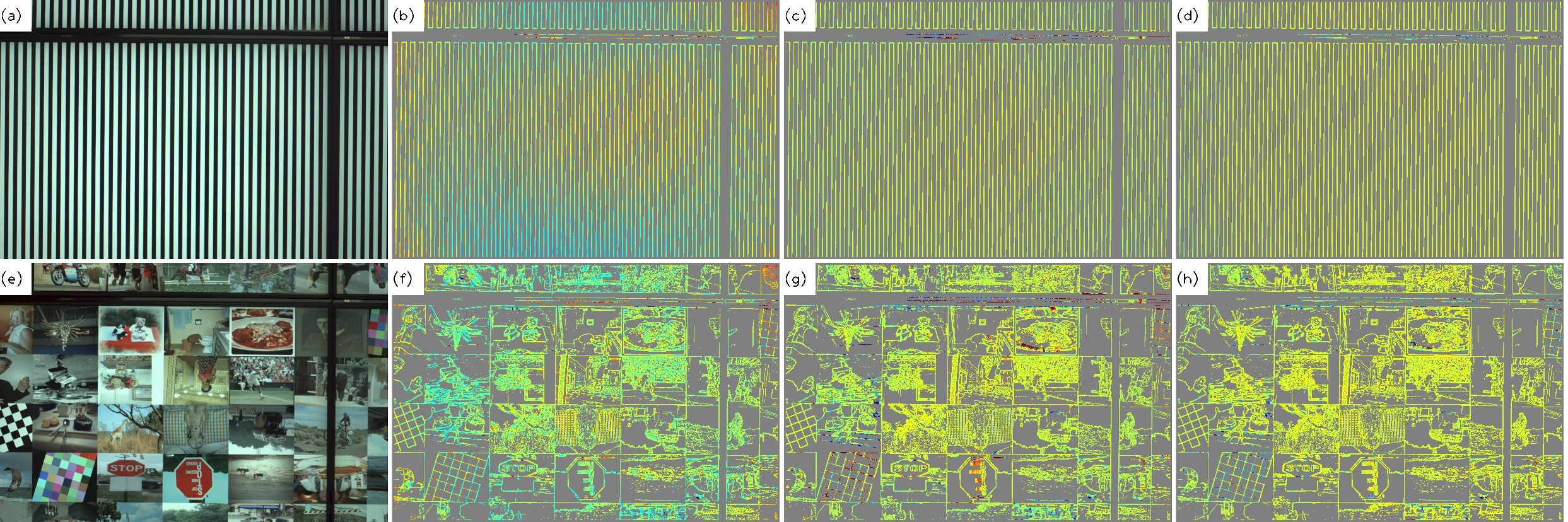}
  \caption{(a), (e) Captured images. (b), (f) DDN without positional branch. (c), (g) DDN without color branch. (d), (h) DDN(ours). }
  \label{depth_map_ablation_study}
\end{figure}

\noindent{\bf Effectiveness of positional branch}\secspace
We trained and tested the network with and without the positional branch using images having several sizes and shapes composed of several blocks as shown in \fig{plot_pos}. In a large area, the training becomes hard because of the need to handle the shift-variant PSF. \fig{plot_pos} shows that the accuracy without the positional branch drops quickly as the area becomes large. As shown in \fig{depth_map_ablation_study}b, the distortion by the shift-variant PSF remains. In contrast, with the positional branch, the accuracy keeps high and the above distortion is disappeared in that depth map as shown in \fig{depth_map_ablation_study}d. 

\noindent{\bf Effectiveness of color branch}\secspace
We evaluate the network with and without the color branch with respect to black-and-white and color subjects as shown in \fig{plot_color}. For the black-and-white subject, two networks have the same accuracy. The accuracy drops in the color subject because high saturated color confuses the network. Since the color branch helps the network to discriminate between the defocus blur and high saturated color, the color branch achieves higher accuracy. This is also shown in the depth maps as shown in \fig{depth_map_ablation_study}g and h. 

\begin{figure}
\centering
  \includegraphics[width=10cm]{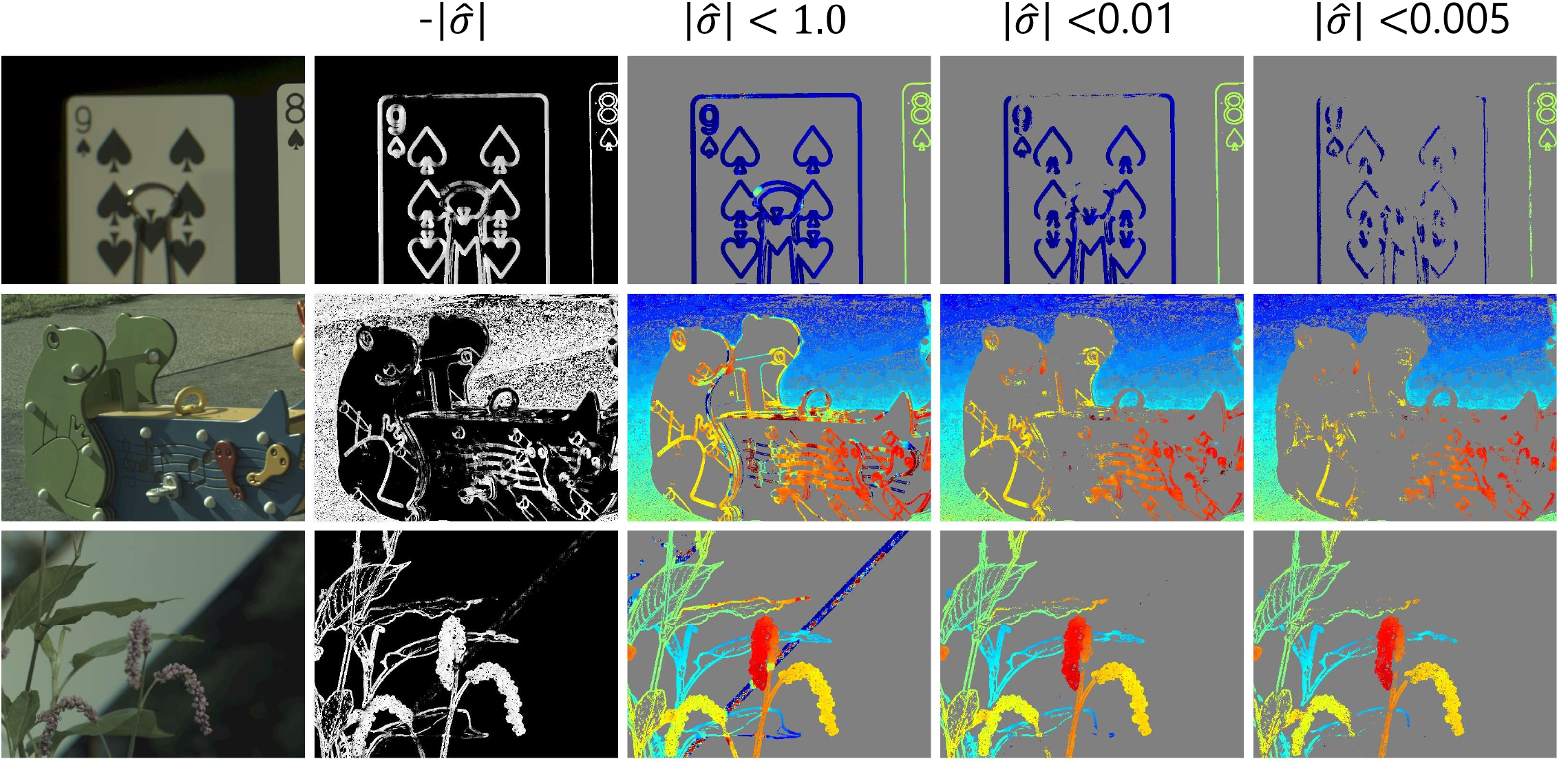}
  \caption{Validation of depth reliability estimation. Column 1: captured images. Column 2: Estimated reliability (darker is lower). Column 3,4,5: Depth maps those unreliable pixels are rejected by three thresholds. }
  \label{depth_uncrt}
\end{figure}

\noindent{\bf Effectiveness of reliability}\secspace
In actual scenes, we verify the effectiveness of the learned reliability. \fig{depth_uncrt} shows several samples of uncertainty. Several types of uncertainty caused depth errors. The reliability prediction can capture the depth errors correctly even though all of them are unseen for the learned network. Threshold $|\hat{\sigma}|<0.01$ shows good balance between the reliable and unreliable regions. 

\subsection{Quantitative and qualitative results}

In the quantitative and qualitative experiments, the DDN is trained by only the indoor data-set. After the training, we changed the focus distance from 1500[mm] to 7000[mm] to apply it to outdoor scenes. We compared our DDN with DfAD~\cite{moriuchi201723} and a stereo camera composed of two prototype CCA cameras with 20cm baseline. Since DfAD utilizes DFD technique to the color channels, the comparison with DfAD includes the one with typical DFDs. Coded-aperture (CA)~\cite{levin2007image} and focal track (FT)~\cite{guo2017focal} have some relations to our method but it is difficult to apply them to our CCA by the following reasons. The image of CCA has fewer zeros on the frequency domain than the requirement of CA. FT is based on the time derivative of defocus blur pairs by the small oscillation lens. It cannot be applicable to CCA due to large differences of blur between the inter-color channels. 

\noindent{\bf Quantitative evaluation}\secspace
We quantitatively evaluated depth errors using our training system. To get stereo depth, we used semi-global matching (SGM)~\cite{hirschmuller2005accurate} implemented by~\cite{opencv}. Although SGM uses strong spatial regularization, DDN and DfAD don't use it. This quantitative evaluation was set to the range from 2000 [mm] to 12000 [mm], which is different from the training. \fig{quantitative_result} shows the error curves over the target distance. The error of DDN is much less than that of DfAD. The error of DDN falls short of the one of the stereo camera. However, the theoretical accuracy of our CCA camera is equivalent to the stereo camera with 1.25cm baseline according to the aperture size. Considering the aperture size, the accuracy of DDN is sufficiently high. 

\begin{figure*}
\centering
\subfigure[Captured image]{\includegraphics[width=3cm,bb=0 0 447 1252]{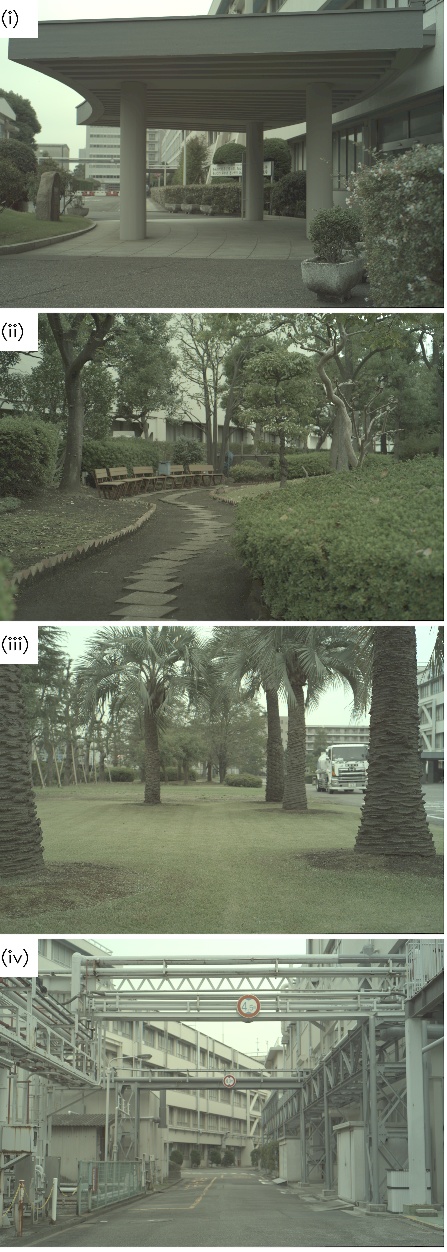}}
\subfigure[Stereo camera]{\includegraphics[width=3cm,bb=0 0 447 1252]{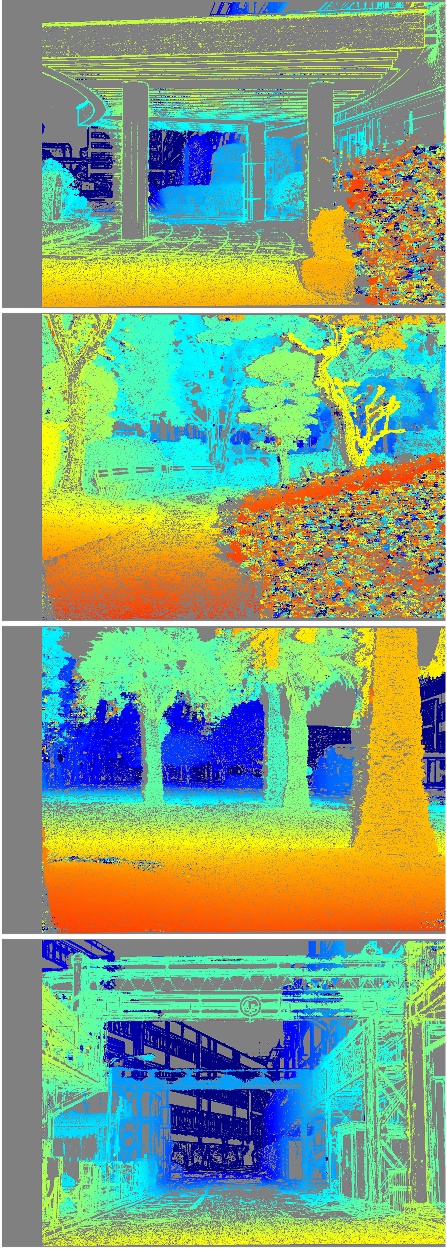}}
\subfigure[DfAD~\cite{moriuchi201723}]{\includegraphics[width=3cm,bb=0 0 447 1252]{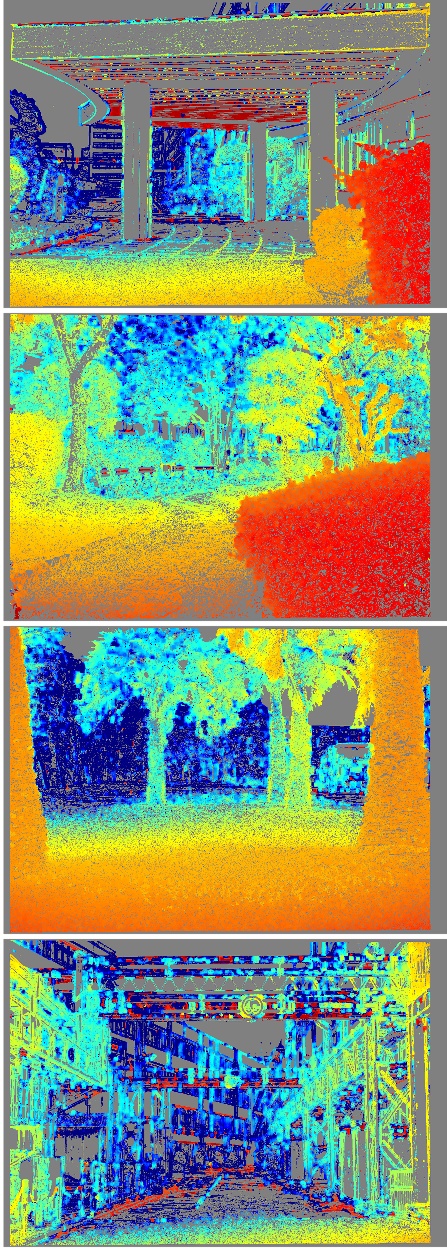}}
\subfigure[DDN(Ours)]{\includegraphics[width=3cm,bb=0 0 447 1252]{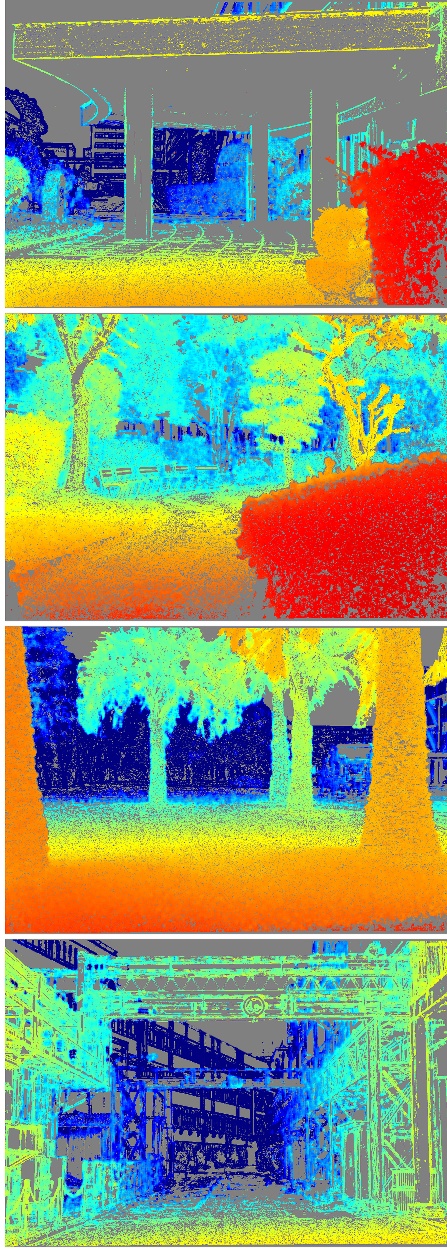}}
\caption{Qualitative comparisons in the case of various outdoor scenes. We compare three methods. We used DDN trained by only the indoor data taken by the training system without any finetuning to the outdoor data. Gray color indicates that there is no depth cue. }
\label{depth_maps}
\end{figure*}

\noindent{\bf Qualitative evaluation}\secspace
We qualitatively evaluated the depth maps in actual outdoor scenes. \fig{depth_maps} shows the qualitative results. Gray color indicates that there is no depth cue. Depth maps by the stereo camera are high resolution at a great distance. However, depth errors often occur at a small distance (within 3[m]) as shown in (i) and (ii). They are caused by occlusion(iii). This is a problem specific to stereo matching. DfAD has insufficient resolution in the distant region((i), (ii), (iii) and (iv)). It also shows several errors caused by horizontal edges((i) and (iv)) and slant edges((iv)). In contrast, DDN gives improved depth maps for the failure cases both of the stereo camera and DfAD. 

\begin{figure*}
\centering
\includegraphics[width=12cm,bb=0 0 3730 616]{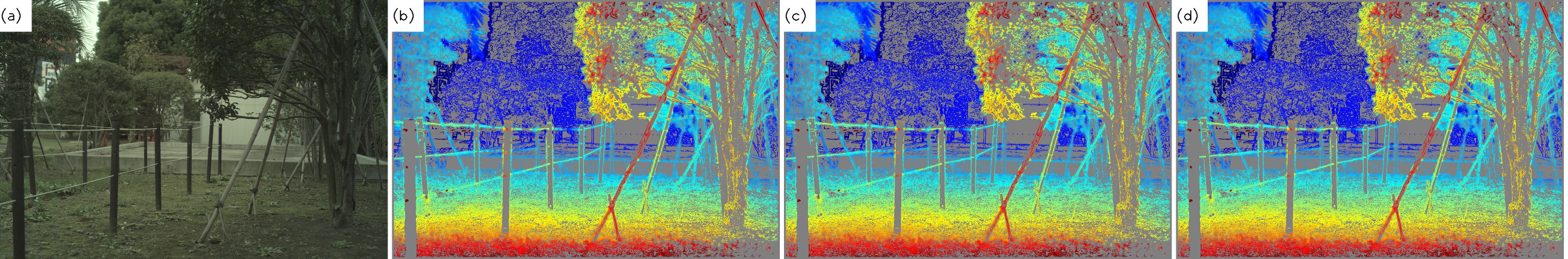}
\caption{(a)Image captured by Lens A for depth estimation. This input image is common for all of the following results. (b) Depth map by DDN trained by Lens A. (c) trained by Lens B (d) trained by Lens C}
\label{other_indivisuals}
\end{figure*}

\begin{figure*}
\centering
  \subfigure[DSLR f=50mm (standard lens)]{\includegraphics[width=4cm,bb=0 0 1855 616]{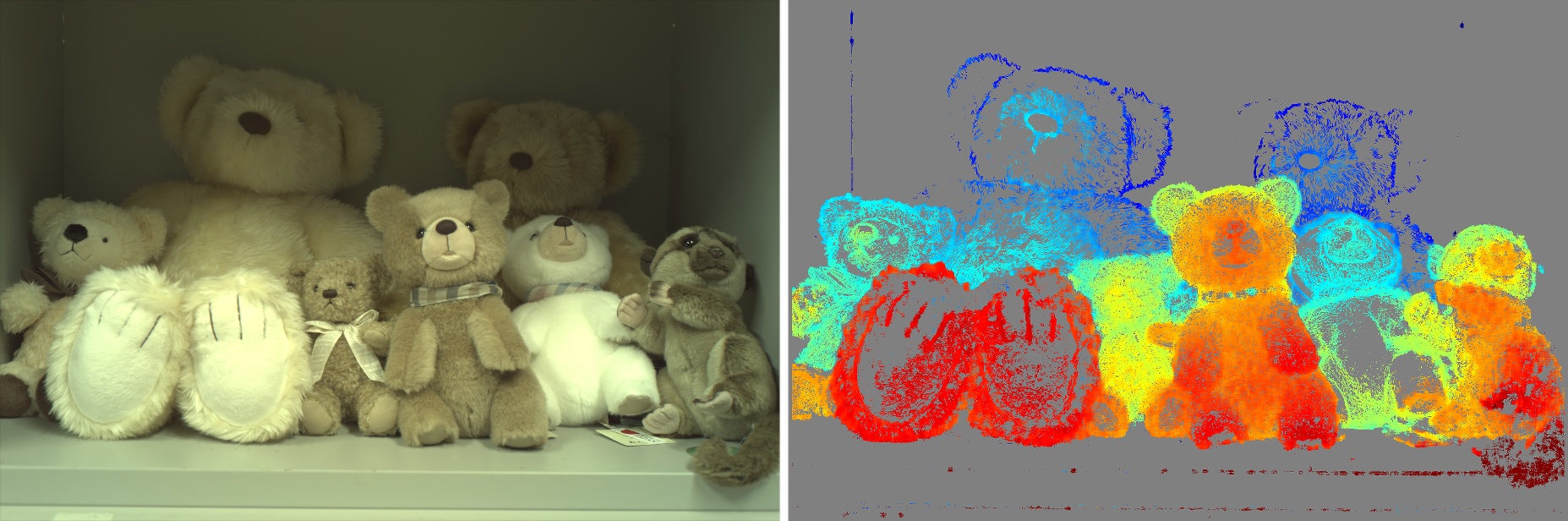}}
  \subfigure[DSLR f=14mm (wide-angle lens)]{\includegraphics[width=4cm,bb=0 0 1855 616]{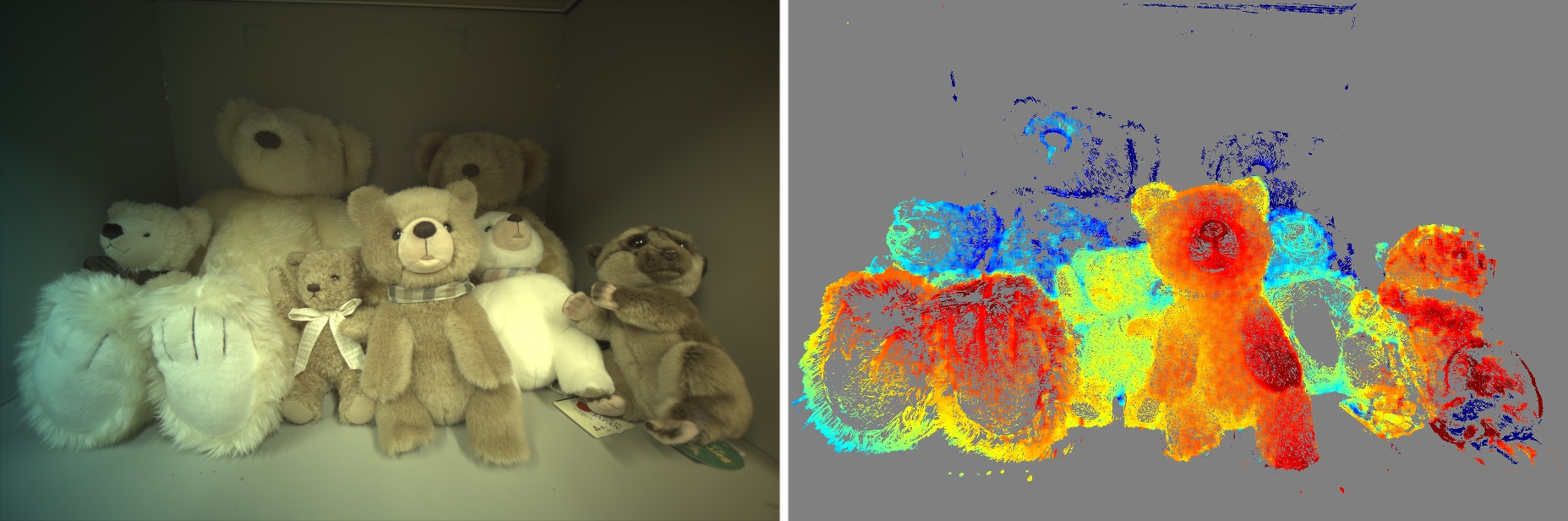}}
  \subfigure[DSLR f=150mm (telephoto lens)]{\includegraphics[width=4cm,bb=0 0 1855 616]{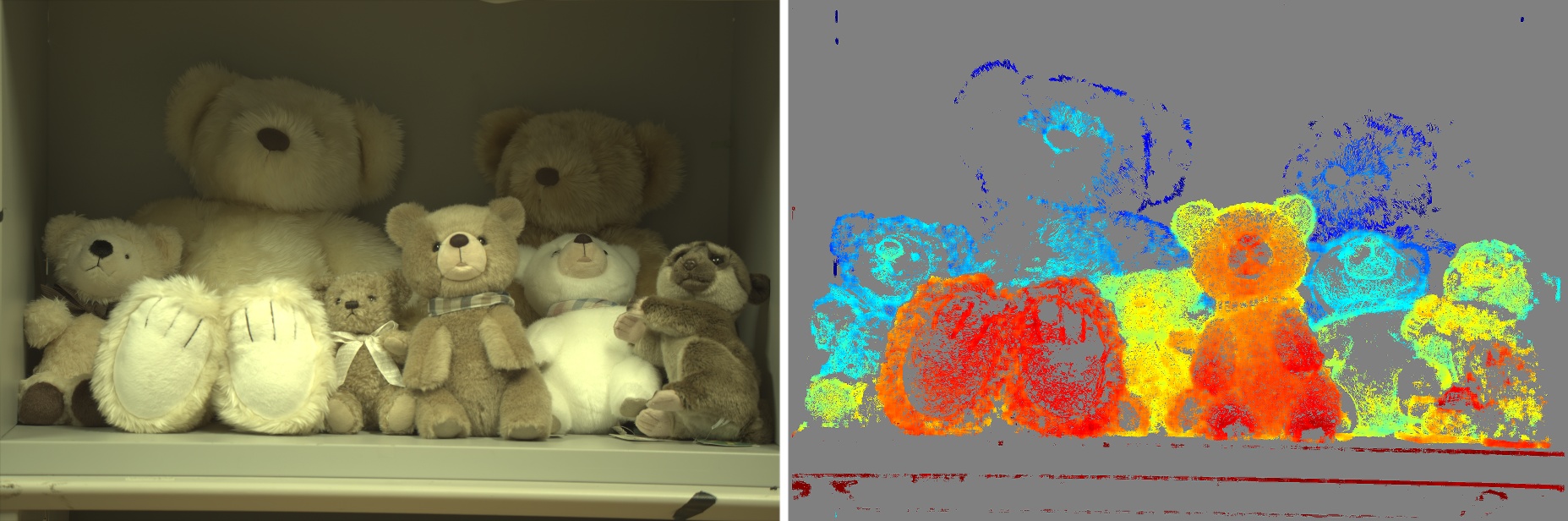}}
  \caption{Depth maps of different focal length. }
  \label{other_lens}
\end{figure*}

\noindent{\bf Robustness against to individual difference and other focal lengths}\secspace
For verification of the robustness against the individual difference, we also trained DDN by Lens B and C (Nikon AI AF Nikkor 50mm f/1.8D). We apply this trained DDNs to the captured image by Lens A (Nikon AI AF Nikkor 50mm f/1.8D) as shown in \fig{other_indivisuals}. There is almost no difference in those depth maps. 
We also apply our method to f=14mm lens (AI AF Nikkor 14mm f/2.8D ED) and f=150mm lens (SP 150-600mm F/5-6.3 Di VC USD G2). As shown in \fig{other_lens}, DDN gives clear depth maps to not only f=50mm lens but also f=14mm lens and f=150mm lens. 

\section{Conclusion}

With a view to realizing the single-shot depth measurement of a monocular camera, we have improved the depth measurement of CCA by using deep learning. We have proposed DDN with a self-attention mechanism to learn lens aberration efficiently. We have also proposed a Bayes L1 loss function to handle the uncertainty more accurately. We have confirmed that DDN showed a great advantage over the baseline in terms of accuracy and, in addition, the accuracy of Bayes L1 loss has been better than L1 loss. The learned reliability has been able to capture the errors caused by uncertainty correctly in spite of unseen outdoor scenes. In terms of quantitative results, the error of DDN was significantly better than DfAD. In terms of qualitative results, DDN was superior to DfAD for various outdoor scenes. 


{\small
\bibliographystyle{ieee}
\bibliography{paper}
}

\end{document}